\documentclass[10pt,twocolumn,letterpaper]{article}

\usepackage{cvpr} 

\usepackage{graphicx}
\usepackage{amsmath}
\usepackage{amssymb}
\usepackage{booktabs}
\usepackage{makecell}
\usepackage{algorithm}
\usepackage{algorithmicx}
\usepackage{algpseudocode}

\newtheorem{theorem}{Theorem}
\newtheorem{definition}{Definition}
\newtheorem{lemma}{Lemma}
\newtheorem{proof}{Proof}

\DeclareMathOperator*{\argmin}{arg\,min}
\usepackage[pagebackref,breaklinks,colorlinks]{hyperref}

\usepackage[capitalize]{cleveref}
\crefname{section}{Sec.}{Secs.}
\Crefname{section}{Section}{Sections}
\Crefname{table}{Table}{Tables}
\crefname{table}{Tab.}{Tabs.}

\begin{document}

\title{Dynamical Isometry based Rigorous Fair Neural Architecture Search}

\author{Jianxiang Luo\textsuperscript{\rm 2} ~ Junyi Hu\textsuperscript{\rm 1,2\thanks{Corresponding author.}} ~ Tianji Pang\textsuperscript{\rm 2} ~Weihao Huang\textsuperscript{\rm 1,2} ~ Chuang Liu\textsuperscript{\rm 2,3}\\
\textsuperscript{\rm 1}Tsinghua University
~ ~ \textsuperscript{\rm 2}Glasssix Technology (Beijing) Group Co., Ltd\\ 
\textsuperscript{\rm 3}Northwestern Polytechnical University
\\
\tt\small \{jianxiangluo,inlmouse,tianjipang,weihaohuang,andyliu\}@glasssix.com    
}

\maketitle

\begin{abstract}
   Recently, the weight-sharing technique has significantly speeded up the training and evaluation procedure of neural architecture search. However, most existing weight-sharing strategies are solely based on experience or observation, which makes the searching results lack interpretability and rationality. In addition, due to the negligence of fairness, current methods are prone to make misjudgments in module evaluation. To address these problems, we propose a novel neural architecture search algorithm based on dynamical isometry. We use the fix point analysis method in the mean field theory to analyze the dynamics behavior in the steady state random neural network, and how dynamic isometry guarantees the fairness of weight-sharing based NAS. Meanwhile, we prove that our module selection strategy is rigorous fair by estimating the generalization error of all modules with well-conditioned Jacobian. Extensive experiments show that, with the same size, the architecture searched by the proposed method can achieve state-of-the-art top-1 validation accuracy on ImageNet classification. In addition, we demonstrate that our method is able to achieve better and more stable training performance without loss of generality.
\end{abstract}

\section{Introduction}
\label{sec:intro}

Neural architecture search (NAS) is a widely used machine learning technology, which automates the design of neural network architecture to find the best model architecture for given tasks. Though the recently proposed weight-sharing strategy helped the traditional NAS methods avoid the burden of training massive neural network architectures from scratch \cite{zoph2016neural,real2017large,eriksson2021latency} and has significantly improved the computational efficiency of current NAS algorithms \cite{liu2018darts,bender2018understanding,xu2019pc,xie2018snas,zhao2021few}, however, the weight-sharing strategy makes the parameters of each candidate subnet highly coupled. This makes subnet candidates hard to obtain actual independent evaluations, leading to insufficient results.

To solve these two problems, Boyu Chen et, al proposed the BNNAS\cite{chen2021bn} algorithm which used the weights of the batch normalization (BN)\cite{guo2020single, chu2021fairnas}layer, named as BN-based indicator, to evaluate the importance of the subnets. During supernet pre-training of BNNAS, only the weights of the BN layer are updated with gradients, while the rest of the random parameters are frozen. As the result, the fixed random parameters and unfixed indicators are successfully decoupled. Though using the BN-based indicator as the subnet performance criterion is inspiring, however, this subnet selecting criterion is purely empirical and has no mathematical guarantees. In addition, a reasonable gradient descent method will affect the results of the architecture search\cite{santra2021gradient}. So it is obviously that random parameters initialization is unable to ensure that the gradients (signal) received by each candidate subnet in each layer are equivalent during the training procedure. As the network goes deeper, this issue becomes progressively severe, affecting the fair evaluation of all BN-based indicators. Furthermore, we discovered that the magnitudes of the BN-based indicators in different layers are not comparable in the module selecting strategies of BNNAS. It is not reasonable to rank the whole performance of the subnets by directly summing the BN scores of each module.

In order to quantify the dynamics behavior of randomly initialized supernet during pretraining, we need to analyze the dynamics of random initialized neural networks. Noticing, the mean field theory (MFT) has been used to establish a theoretical understanding of neural networks with random parameters and quantitatively portrayed the “average” dynamics of signal propagation inside of themselves\cite{poole2016exponential, schoenholz2016deep, schoenholz2017correspondence, pennington2017resurrecting, xiao2018dynamical}. MFT and extensive experiments all showed that networks can be trained most efficiently and stably when their input-output Jacobian of every layer achieves dynamical isometry(orthogonal), namely the property that the entire distribution of singular values of the Jacobian matrix is close to 1\cite{poole2016exponential,schoenholz2016deep}. The well-conditioned Jacobian ensures stable propagation while avoiding the vanishing or exploding of the signal. In the NAS algorithm, by initializing each module of the supernet to achieve dynamic isometry, the input signal of the network can be equally propagated to any place of the supernet, which is able to reflect the architecture information in neural networks as much as possible. 

In the present work, we continue the line of most one-shot and weight-sharing based NAS techniques and proposed a fairer and more efficient approach for neural architecture search. Specifically, we obtained the dynamical isometry module by triangular decomposing the randomly initialized weights of gaussian distribution. This network weights initialization strategy ensures that each candidate module is dynamical isometry while remaining frozen during supernet pretraining. The input signal can be equally propagated to each and every module in the search space horizontally and vertically. Following these principles, the BN-based indicators are pitched into the bottom of each module in the subnets as the performance indicators of those modules with different structures. To deal with the above-mentioned module selecting dilemma, we select the module with largest BN-based indicator as the target module of each layer. At last, we present rigorous proof of the feasibility of using the parameters of the BN layer as subnets evaluation criterion. Extensively experiment showed, with the same size, the architecture searched by the proposed method can achieve state-of-the-art top-1 validation accuracy on ImageNet classification. In addition, we demonstrated the proposed method is able to achieve better and more stable training performance without loss of generality.

The contributions of this work can be summarized as follows:
\begin{enumerate}
	\item We designed an initialization method for NAS algorithms which guarantees the equivalent inputs for both shallow and deep modules and ensures the fairness of search evaluation.
	\item We give a mathematical proof that the value of the BN layer’s parameters is able to reveal the signal propagation capability of the candidate modules, which for the first time shows the interpretability and rationality of the BN-based indicator theoretically.
	\item We point out a new module selection strategy which fixed the problem of selecting the module by the BN-based indicator’s numerical scale across layers is uneven.
\end{enumerate}

\section{Related works}

\subsection{Weight-sharing NAS}

We denote the search space of a weight-sharing based NAS algorithm as $\mathcal{A}$. The entire supernet can be denoted as $\mathcal{N} (\mathcal{A},\mathbf{W})$, where $\mathbf{W}$ denotes the parameter weights of the supernet. Then, all parameter weights of the supernet can be jointly optimized as follows:

\begin{equation}
	\mathbf{W}_{\mathcal{A}} = \argmin_{\mathbf{W}} \mathcal{L} ( \mathcal{N} ( \mathcal{A}, \mathbf{W} ) ).
\end{equation}

Single path one-shot NAS\cite{guo2020single} avoids the coupling of weights in the joint optimization by selecting a uniform path of candidate modules from each layer to form a subnet, and trains each subnet’s path individually. The component module of the subnet is $a \subseteq \mathcal{A}$ with parameters $\mathbf{W}_a$. And on this basis, FairNAS\cite{chu2021fairnas} analyzes the subnet sampling strategy and proposes a "strict fairness" method. We use the notation $a \sim \Phi(\mathcal{A})$ to denote this sampling strategy:

\begin{equation}
	\mathbf{W}_{\mathcal{A}} = \argmin_{\mathbf{W}_a} \mathbb{E}_{a \sim \Phi(\mathcal{A})} \left[ \mathcal{L} ( \mathcal{N} ( a, \mathbf{W}_a ) ) \right].
\end{equation}

After the supernet being pre-trained, the weight-sharing NAS method evaluates the accuracy of each subnet on the validation set one by one or with some strategies to obtain the best subnet structure. This selected subnet is generally retrained based on the training set to obtain the final output of the NAS algorithm.

BNNAS\cite{chen2021bn} is inspired by \cite{frankle2020training}, which studies the role of the Batch Normalization layer in the network’s forward propagation. They heuristically came up with the idea of making full use of the BN layer in NAS..This method regards the BN layers’ learnable weights as the evaluation of the candidate module and names those layers as the BN indicator. During the training of the supernet, only the parameters of the BN layers are back-propagated with gradients while the remaining parameters in the network are frozen. After the supernet being trained, BNNAS uses the value of the optimized BN indicator’s weight as the evaluation criterion for the current module. An evolutionary algorithm is used to meritively select modules from the candidate set and combines the selected modules into a final subnet.

Though this idea improves the searching speed of the NAS algorithm, however, it ignores the fairness in training and search procedure. Since the weights of individual modules in the supernet are randomly initialized and frozen during supernet training, the elusive parameters can affect the structural expressiveness of modules. It is also difficult for modules in deep layers to obtain ``equivalent" input feature maps for fair cross-sectional comparison. Inspired by mean-field theory, we find that signal propagation in the network can solve the above problem when dynamical isometry is achieved. We can also explain the plausibility of the BN-indicator in a neural network under the mean field.

\subsection{Dynamical Isometry}

The dynamics of a neural network as the signals propagate through is a classical research subject\cite{xiao2018dynamical, pennington2017resurrecting, speicher1994multiplicative, miyato2018spectral}. A. M. Saxe et. al regard the learning dynamics on the weight space as a set of nonlinear differential equations\cite{saxe2013exact}, which describes the dynamics as a function between the inputs and outputs of the network. According to this equation, the Jacobian matrix of the input-output of the network can be quantified. 

Consider a neural network with $L$ layers, which is composed of a series of linear and nonlinear functions. $\mathbf{W}_{l}$ and $\mathbf{b}_{l}$ denote the weight matrices and bias vectors of the linear function in the $l$-th layer, with $l=1,\cdots, L$. The activation function is $\sigma(\cdot)$. The forward propagation of the network is:

\begin{equation}
	\mathbf{h}_{(l+1)} = \sigma \left( \mathbf{W}_{l} \cdot \mathbf{h}_{(l)} + \mathbf{b}_{l} \right),
\end{equation}
where $\mathbf{h}_{l}$ represents the state of the signal at different layers. We denote the input-output Jacobian matrix of the network by $\mathbf{J}$, and the non-linear part of the network is $\mathbf{D}$. $\mathbf{J}$ can be used to describe the relationship between the input and the output signal of any layer of the network:
\begin{equation} 
	\mathbf{J}= \frac{\partial \mathbf{h}_{l}}{\partial \mathbf{h}_{0}} = \prod _{j=1}^{l} \mathbf{D}_{j} \mathbf{W}_{j}.
\end{equation}

\textbf{Order-to-chaos phase transition and Fixed point analysis}. Singular values $\chi$ of the Jacobian matrix affect the signal propagation in terms of network depth. \cite{poole2016exponential} and  \cite{schoenholz2016deep} studied the order-to-chaos phase transition of the network through $\chi$:
\begin{equation}
	\chi = \phi \left(\left( \mathbf{D}\mathbf{W}\right) ^{T} \mathbf{D}\mathbf{W}\right), 
\end{equation}
where $\phi$ is the function that computes the expectation. The value of $\chi$ is influenced by the initialization of the network weights. Only $\chi=1$ can avoid gradient disappearance or gradient explosion, that is, chaos phase or order phase.

The fixed point of the signal in the process of network propagation, which is related to the variance change of the signal. (\cite{poole2016exponential}) study the variance of the input-output of each module in the network. Denote the variance of the module’s weight is  $v_{\mathbf{W}}$, the variance of the bias is $v_{\mathbf{b}}$. Then when the input signal satisfies a Gaussian distribution with mean $0$ and variance $v_h$, the variance of the output of this signal after the module is:

\begin{equation} 
	v_{\mathbf{h}_{l}} = v_{\mathbf{W}} \int \mathcal{D}h \sigma  
	\left( \sqrt{v_{\mathbf{h}_{l-1}}}h\right)^2 + v _{\mathbf{b}},
\end{equation}
where $\mathcal{D}h = \frac{d h}{\sqrt{2 \pi}} e^{-\frac{h^2}{2}}$ denotes the standard Gaussian measure. When the propagation of the signal is at a fixed point, the variance of the input and output vectors of any module in the network is the same. We say that the subnets on the fixed point can achieve dynamic isometry:

\begin{equation}
	v_{\mathbf{h}_*} = v_{\mathbf{W}} \int \mathcal{D}h \sigma  
	\left( \sqrt{v_{\mathbf{h}_*}}h\right)^2 + v _{\mathbf{b}}.
\end{equation}

\textbf{Gaussian weight initialization}. \cite{pennington2017resurrecting} demonstrates the difficulty of using a Gaussian distribution to initialize the parameters of a network model to achieve dynamic isometry. They derived the Stieltjes transform from the limiting spectral density of the matrix $\mathbf{J}\mathbf{J}^T$, and generate the moment of it $M_{\mathbf{J}\mathbf{J}^T}$. The first two moments denote $m_1$ and $m_2$, and have:

\begin{equation}
	\begin{aligned}
		m_1&=(v_{\mathbf{W}} p(\mathbf{h}_*) ) ^ L \\
		 m_2 &= (v_{\mathbf{W}} p(\mathbf{h}_*)) ^{2L} (L + p(\mathbf{h}_*)) / p(\mathbf{h}_*).
	\end{aligned}	
\end{equation}
where $p(\mathbf{h}_*)$ denotes the probability that the current module is a linear function, which can also be expressed as the proportion of neural operating in the linear regime. Both first and second moments increase exponentially with the number of layers. When network are at the  fixed points, that is $m_1 \approx 1$, the variance is $\frac{L}{p( \mathbf{h} _* )}$ which continues to increase with the number of layers. Therefore, for any neural network, Gaussian weight initialization cannot be selected to avoid this growth.

\section{Method}
\subsection{Dynamic Isometry in NAS}

In order to evaluate the candidate modules of the neural network in the supernet, we refine the input-output Jacobian matrices of the entire network to individual modules. Each subnet in the search space is composed of the subposition of independent continuously modules. Our contribution is that when candidate modules at the same level are compared horizontally, the dynamics of the modules remain the same, which provides absolute fairness for the subsequent evaluation mechanism.

\subsubsection{Candidate module dynamic isometric}
We denote $\mathbf{J} _{l,m}$ as the the input-output Jacobian matrix of the $m$-th candidate module in the $l$-th layer in the network, where $l = 1,\cdots,L$ and $m = 1,\cdots,M$. The signal in the network add the index of the candidate module, $\mathbf{h}_{l,m}$.

\begin{equation}
	\mathbf{J}_{l,m} = \frac{\partial \mathbf{h}_{l,m}}{\partial \mathbf{h}_{l-1,m}}.
\end{equation}

To explore the dynamic isometry of candidate modules according to $\mathbf{J}_{l,m}$, we denote $\lambda _i$ as the $i$-th eigenvalue of $\mathbf{J}_{l,m} \mathbf{J}_{l,m}^T$. We have $\text{tr}(\mathbf{J}_{l,m}\mathbf{J}_{l,m}^T) = \frac{1}{w} \sum_{i=1}^{w} \lambda_i$ and assume the eigenvalues of this matrix are independent of each other. The parameter $w$ can reflect the width of the network to some extent through the Jacobian matrix. The variance of $\text{tr} (\mathbf{J} _{l,m} \mathbf{J} _{l,m}^T)$ can be given by

\begin{equation}  \label{equation_second_moment}
	\begin{aligned}
		&D \left[ \text{tr}\left( \mathbf{J}_{l,m} \mathbf{J}_{l,m}^T\right) \right] \\
		&= \frac{1}{w} \sum_{i=1}^{w} \mathbb{E} \left[\lambda _i^2\right] - \mathbb{E}^2\left[ \lambda_i\right] \\
		&= \phi \left(\left( \mathbf{J}_{l,m} \mathbf{J}_{l,m}^T\right) ^2\right) - \phi ^2 \left(\mathbf{J}_{l,m} \mathbf{J}_{l,m}^T\right) \\
		&:= \varphi \left( \mathbf{J}_{l,m} \mathbf{J}_{l,m}^T\right).
	\end{aligned}
\end{equation}

According to the satisfied conditions of dynamic isometric and the above formula, we define that the condition under which the candidate module can achieve dynamic isometric is 

\begin{equation} \label{equation_isometry_condition}
	\begin{aligned}
		\phi \left(\mathbf{J}_{l,m} \mathbf{J}_{l,m}^T\right) & \approx 1, \\
		\varphi \left(\mathbf{J}_{l,m} \mathbf{J}_{l,m}^T \right) & \approx 0.
	\end{aligned}
\end{equation}

We can use the moment generating function to prove this setting by referring to the \emph{Tanh} activation and Orthogonal weights of corresponding literature. In fact $\phi \left(\mathbf{J}_{l,m} \mathbf{J} _{l,m} ^T\right)$ and $\varphi \left(\mathbf{J} _{l,m} \mathbf{J} _{l,m} ^T \right)$ can be considered as the first two moments of $\mathbf{J} _{l,m}$. With this setting, we can ensure that the expectation of the singular values of each candidate module's input-output Jacobian matrices is close to 1. It is worth mentioning that, according to equation~\ref{equation_second_moment}, we find when the network width of each candidate module is large enough, this expectation can be equal to 1.

The input $\mathbf{h} _0$ is the same for all subnets in the search space. When we guarantee each module to achieves dynamic isometry, as the number of layers increases, the dynamics of the signal  passed to the subsequent candidate modules remain stable. The only thing that affects the output of the model is the structure of the network itself, which is the original intention of our neural network architecture search.

\subsubsection{Orthogonal Initialization} \label{section_orth_init}

While keeping the supernet weights frozen, a well-designed initialization method is needed in order to achieve dynamic isometrics of the network. If a candidate module implements orthogonal weights, for all the parameters of the linear function in that module we have $\mathbf{W} \mathbf{W}^T = \mathbf{I}$. We want the Jacobian of this module $\mathbf{J} _{l,m} := \mathbf{D} _{l,m}  \mathbf{W} _{l,m}$ to satisfy

\begin{equation} 
	\phi \left( \mathbf{J} _{l,m} \mathbf{J} _{l,m}^T \right) = 1.
\end{equation}

For different activation functions, suitable orthogonal initialization can achieve $\phi (\mathbf{JJ} ^T)=1$. However, in order to achieve dynamic isometry according to the equation~\ref{equation_isometry_condition}, we need to satisfy$\varphi(\mathbf{JJ}^T)=0$. The activation function $Tanh$ is defined as $f(x)=\frac{2}{1+e^{-2x}} -1$, and for derivation $f ^{\prime}(x) = 1 - f ^2(x)$. According to the properties of  $Tanh$ we have $\forall x\in \mathbb{R} ,|tanh(x)|/|x|<1$, therefore, when the number of layers is large enough, almost the function will concentrate around 0. Then the Taylor expansion reduces to $f(x) \approx f(0) + f ^{\prime} (0) x=x$, The above Taylor formula is similar to the identity matrix, approximately, we have $\phi(\mathbf{JJ} ^T)=1$ and $\varphi(\mathbf{JJ} ^T)=0$.

After satisfying the dynamic isometry condition of equation~\ref{equation_isometry_condition}, we ensure that the input and output signal of the same module have the identical mean and variance.

\subsection{How Dynamic Isometry Guarantees Fair Evaluation}

\begin{figure}[h]
	\centering
	\includegraphics[width=\linewidth]{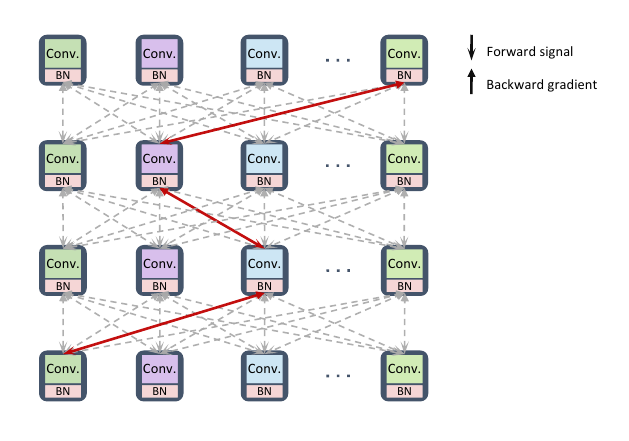}
	\caption{The connection between layers is determined by the training strategy, and it is difficult to determine the structure of the subnet from the beginning. The input of the candidate module of the latter layer may come from all modules from the previous layer}
	\label{figure_supernet_layer_connection}
\end{figure}

The complex search space is organized into a regular directed acyclic figure~\ref{figure_supernet_layer_connection}, and the evaluation BN indicators are embedded in each module as part of the network. Under the premise of choosing a topological path, the candidate modules in the first layer network are lucky, they accept the same input of the supernet $\mathbf{h}_{0,m} \in \mathbb{R} ^{n \times n \times d}$. However, it is difficult for candidate modules in subsequent layers to have the same input, each time they accept different outputs from different candidate modules in the upper layer.

In this chapter, we explain the grounds that our search algorithm is absolutely fair under the premise that every candidate module in the search space is dynamical isometry and how does our initialization method guarantee that the outputs or inputs of candidate modules in the same layer are nearly equal.

Assuming that the initialization parameters of each module have made it meet the conditions of dynamic isometry, the output of modules can be denoted as $\mathbf{h}_{l,m}$. The input-output Jacobian matrix of each module can be denoted as $\mathbf{J}_{l,m}$.

The linear part in the input-output Jacobian matrix $\mathbf{J}_{l,m}$ is denoted as $f_{l,m}$. In this paper, the linear function is the convolution operation. Let $F_{p} \in \mathbb{R}^{r \times r}$ denotes the filters in $f_{l,m}$, the number of which is $N$, where $p$ denote the index. The remaining activation function methods we denote as $\sigma(\cdot)$, we concentrate on the $tanh$ function.

For the Batch normalization layers in the block and our BN-based indicator, we state their representation

\begin{equation}
	\begin{aligned}
		\mathbf{h}_{l,m} ^{bo}= \gamma \cdot \frac{\mathbf{h}_{l,m} ^{bi}-\mathbb{E}[\mathbf{h} _{l,m} ^{bi}]}
		{\sqrt{\text{Var} ^2[\mathbf{h}_{l,m} ^{bi}] - \varepsilon}} + \beta  , 
	\end{aligned}
\end{equation}
where $\mathbf{h}_{l,m} ^{bi}$ and $\mathbf{h}_{l,m} ^{bo}$ represent the input and output of the Batch normalization in each module, respectively. The weight $\gamma$ and bias $\beta$ are learnable parameters in BN layers to affine the normalized features. $\varepsilon$ is a positive number. We set $\mathbb{E}[\mathbf{h} _{l,m}]=0$ and the bias is $0$. In addition, $v_h$ denote the variance of $\mathbf{h} _{l,m}$. Then the representation of the BN layer function is simplified to

\begin{equation} 
	\text{B}({h}_{l,m}) := \frac{\gamma}{\sqrt{v_h-\varepsilon}} {h}_{l,m}.
\end{equation}

Since different measures $f_{l,m}$ are independently and identically distributed, we take one of the modules for analysis. Ideally, the changes that occur in the signal after passing through the convolution module should be caused by the structure of the module. However, even modules with well optimised parameters may bring biases into the signal's feature extraction. This problem is exacerbated when randomised parameters are used, and biases accumulate cascading through the supernet training. 

We investigated that the impact caused by such randomised parameters is very limited in the case where the modules are dynamically isometry. Considering two random tensors as inputs to the same module, we analyse the change in the output distance of the convolution module for the two random variables. We show that the expected distance between the output of the volume tensor and the actual distance is controlled probabilistically when the module satisfies dynamic isometry.In the process of filter inner product, the part of the signal $\mathbf{h}_{l,m}$ is defined as $[x]_{i,j}^r \in \mathbb{R} ^{r \times r \times d}$, and $i,j \in \mathbf{Z}_n$ indicate location. For another input $\mathbf{h}_{l,m}^{\prime}$ using the same method to do the split and get $[y]_{i,j}^r \in \mathbb{R} ^{r \times r \times d}$.

\begin{theorem} \label{theorem1}
	Let $*$ denotes the cyclic convolution operation, $\mathbf{h},\mathbf{h}^{\prime} \in \mathbb{R} ^{n\times n\times d}$ be inputs to a filter $F_p \in \mathbb{R}^{r \times r \times d}$, with $r<n$. The BN and activation function are expressed as $B(\cdot)$ and $\sigma(\cdot)$ respectively. All of $F_p$ denote $F$, which are independent and identically distributed, and orthogonal tensors decomposed from Gaussian distributed tensors with variance $v^2$.
	
	The expectation output of the inner product of two tensors is
	
	\begin{equation}
		\mathbb{E} \left[ \left\langle B\left( \sigma\left( F * \mathbf{h}\right)\right), B\left( \sigma\left( F * \mathbf{h}^{\prime}\right)\right)\right\rangle \right].
	\end{equation}
	Moreover, we denote the variance of the two inputs as $v_{h}^{\prime} = \left\lVert (v_h - \varepsilon) ^{-\frac{1}{2}} \right\rVert $, $v_{h^{\prime}}^{\prime} = \left\lVert (v_{h^{\prime}} - \varepsilon) ^{-\frac{1}{2}}\right\rVert $ and have
	
	\begin{equation}
		R := \max \left\{ \max_{i,j \in \mathbb{Z}_n} \left\lVert [x]_{ij}^r\right\rVert v_h^{\prime},  
		\max_{i,j \in \mathbb{Z}_n} \left\lVert [y]_{ij}^r\right\rVert v_{h^{\prime}}^{\prime} \right\}
	\end{equation}
	
	While the activation function satisfies the Lipschitz condition and the Lipschitz constant is $L$, there is at most $\delta > 0$ probability such that the difference between the expected and the actual output satisfies:
	\begin{equation}\label{equation_prob_distance}
		\begin{aligned}
			\mathbb{P}\left[ \left| \frac{1}{N} \sum_{p=1}^{N} \left\langle B\left( \sigma\left( F_p * \mathbf{h} \right) \right) , B \left( \sigma \left( F_p * \mathbf{h} ^{\prime} \right) \right) \right\rangle \right.\right. \\
			\left.\left. - \mathbb{E} \left[ \left\langle B \left( \sigma\left( F * \mathbf{h}\right)\right), B\left( \sigma\left( F * \mathbf{h} ^{\prime} \right) \right) \right\rangle \right] \right| \geq \epsilon \right] \leq \delta \\
		\end{aligned}
	\end{equation}
	for $\delta=2n^2 \exp \left\{-min\left( K^2,K\right)cN\right\}$, where $c>0$ is an absolute constant and $K = \epsilon/D \left\lVert \gamma\right\rVert  ^2 v^2 L^2 R^2 n^2 $ while $D>0$ is an absolute constant.   
	
\end{theorem}

Since the network parameters are untrained, it is almost impossible to achieve the expected state of the candidate module. We assume that $\epsilon$ in equation~\ref{equation_prob_distance} is always present and fixed. Then by adjusting $\delta$ so that the distance of change of the output tensor approximates the theoretical distance of change of the output with maximum probability, is to make $\delta$ as small as possible. Then after freezing the weights of the candidate modules with implementing dynamic isometry, only the parameter $\gamma$ of the BN layer is the only variable.

As the value of $\gamma$ increases, the closer the actual distance between any two outputs is to their desired distance.That is, the more the output signal changes to reflect the structural information of the module itself, avoiding interference and deviations caused by the randomness of the parameters. A larger value of $\gamma$ reflects a fairer neural network signaling, the change in signal is more responsive to the structural information of the module.

With the implementation of dynamic isometry, the randomness of the initialisation parameters of the candidate modules has a limited impact on signal propagation, especially in the case of fixed weights.In addition optimising the parameters of the BN layer is beneficial for NAS fairness improvement, the value size of $\gamma$ reduces the side effects of fixed parameters and promotes fair NAS evaluation.

\begin{figure}[h]
	\centering
	\includegraphics[width=\linewidth]{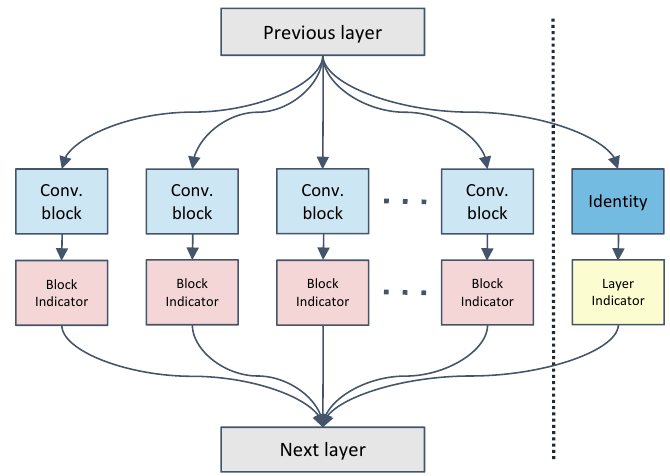}
	\caption{The sketch map of BN-based layer indicator in a candidate module. The Conv Block refers to a series of operations including various convolutional layers, activation functions, ordinary BN layers, etc.}
	\label{figure_BN_layer_indicator}
\end{figure}

\subsection{Fairness Improvements for NAS}

In order to facilitate the distinction, we call BNNAS's BN-indicator as BN-based block indicator, which measures the importance of the block. Its numerical value determines whether the current module is outstanding.

\begin{algorithm}
	\label{algorithm1}
	\renewcommand{\algorithmicrequire}{\textbf{Input:}}
	\renewcommand{\algorithmicensure}{\textbf{Output:}}
	\caption{Subnet Selection}
	\begin{algorithmic}
		\Require Supernet $\mathcal{N}$ with search space $\mathcal{A}$, the layer of $\mathcal{N}$ is $L$ and $\mathcal{N}^l$ denote every layer, the number of blocks in every layer is $M$. BN-based layer indicators $\mathbf{U}^{l}$.
		\Ensure the Selected architecture $\varGamma $ of subnet
		\State Set a list $\varGamma $ \;
		\For {$l$ \textbf{in} $L$}
		\If {$\mathcal{N}^l$ is reduction layer}
		\State Randomly select a $\mathcal{N}^{l,m}$ that has not been selected recently from the $M$ candidate modules of the current layer
		\EndIf
		\If {$\mathcal{N}^l$ is normal layer}
		\State Randomly select a $\mathcal{N}^{l,m}$ that has not been selected recently from the $M$ candidate modules or $\mathbf{U}^{l}$
		\EndIf
		\State $\varGamma \leftarrow \varGamma.\text{append}(\mathcal{N}^{l,m})$ 
		\EndFor
	\end{algorithmic}
\end{algorithm}

If only to find one optimal subnet, we need to select the highest-rated candidate module in each layer of the network. However, due to some circumstances, when we want to search several good candidate subnets, simply adding up the weight averages and sorting is unconvincing. Since the BN-based block indicators are difficult to compare vertically, it is hard for us to obtain the module importance relationship between layers. Therefore, we add a BN-based layer indicator to every layer to measure the importance of the current layer in the entire network.
\begin{algorithm} [h]
	\label{algorithm2}
	\renewcommand{\algorithmicrequire}{\textbf{Input:}}
	\renewcommand{\algorithmicensure}{\textbf{Output:}}
	\caption{NAS algorithm}
	\begin{algorithmic}
		\Require Supernet $\mathcal{N}$ with search space $\mathcal{A}$, the layer of $\mathcal{N}$ is $L$ and $\mathcal{N}^l$ denote every layer, the number of blocks in every layer is $M$. BN-based layer indicators $\mathbf{U}^{l}$ with parameters $\theta ^{l}_{u}$; BN-based block indicators $\mathbf{V} ^{l,m}$ with parameters $\theta ^{l,m}_{v}$, other parameters in $\mathcal{N}$ is $\theta$. Train data $\mathcal{D}$, and train epochs is $E$, batchsize denote $B$. Search constraint $C$
		\Ensure Searched $K$ architectures with trained parameters
		\State Orthogonal initialization for parameters $\theta$ in the network $\mathcal{N}$ \Comment{ \textbf{Initialization}}
		\For {$i \leftarrow 0$ \textbf{to} $E$}  \Comment{\textbf{Train phase}}
		\For{$data$ \textbf{in} ($\mathcal{D}$, $B$)} 
		\State $Arch$ $\leftarrow$ Select a Subnet($\mathcal{N}$, $L$) \;
		\State Training $\theta ^{l}_{u}$ and $\theta ^{l,m}_{v}$ in $\mathbf{U} ^{l}$ and $\mathbf{V} ^{l,m}$ separately via $data$, and fix $\theta$\;
		\EndFor
		\EndFor
		\For {$l$ \textbf{in} $L$} \Comment{\textbf{Search phase}}
		\For {$m$ \textbf{in} $M$} 
		\If {$\mathcal{N}$ is reduction layer}
		\State $S ^{l,m} \leftarrow \text{Mean}(\text{Abs}(\theta ^{l,m}_v ))$
		\EndIf
		\If {$\mathcal{N}$ is normal layer}
		\State $S ^{l,m} \leftarrow \text{Mean}(\text{Abs}(\theta ^{l,m}_v)) * \text{Mean} (\text{Abs} (\theta ^{l}_u))$
		\EndIf
		\EndFor   
		\EndFor
		\State Sampling $K$ subnets under the constraints of $C$ according to the rank of the sum of $S ^{l,m}$ in each subnet.
		\For {$k$ \textbf{in} $K$} \Comment{\textbf{Retrain phase}}
		\State Train the selected subnet with $\theta$ and $\theta ^{l,m}_v$ via $\mathcal{D}$
		\EndFor
	\end{algorithmic}
\end{algorithm}
The network layer is divided into reduced layers and normal layers. Considering the indispensability of the reduced layer, we do not evaluate it. We add a parallel identity connection to each layer in the network and a BN layer after the identity as the BN-based layer indicator. 

Figure~\ref{figure_BN_layer_indicator} shows the details of the BN-based layer indicator.The sampling method of the algorithm after adding BN-based layer indicator is shown in algorithm~1. We still use the method of fair sampling based on FairNAS\cite{chu2021fairnas}, so that each candidate module has the same expectation of being selected. The pseudo code of the algorithm for the entire search process is algorithm~2.

In addition, the training strategy based on SPOS does not affect the dynamic isometry of the network when the subnet selects the BN-based layer indicator.

\section{Experiments}

To prove the effectiveness of our parameter initialization method, we apply our weight initialization method to the NAS algorithm with the addition of BN-based layer indicator and set up comparative experiments. For comparative experiments with state-of-the-art methods, our experimental steps are strictly performed under the settings of BNNAS and SPOS\cite{guo2020single}.

\textbf{Dataset}. For the dataset's choice, we choose to use ImageNet\cite{deng2009imagenet}, which contains a train set with over one million samples and a validation set with fifty thousand samples. Only the training set needs to be used in the training and search of the supernet, and the validation set will not be obtained until the found subnet is retrained.

\textbf{Search Space}. The search space follows the same settings as BNNAS and SPOS. Firstly, the search space of the first set of experiments is mainly composed of MobileNetV2 blocks\cite{sandler2018mobilenetv2} with the kernel size in $\{3,5,7\}$ and internal expansion rate in $\{3,6\}$ for candidate modules. Different kernel sizes and internal expansion rates can combine to six candidate modules for each layer in the supernet to choose from. The entire supernet totally has 20 layers, including reduced layers and normal layers. In addition, for another set of experiments, we set the ShuffleNetV2 block\cite{ma2018shufflenet} as the subject of the candidate module. The number of layers of the network is set to 20, and the candidate modules of each layer are composed of ShuffleNetV2 block with kernel sizes of $\{3,5,7\}$, and a ShuffleNet Xception block\cite{chollet2017xception}. The size of the search space for these two sets of experiments is $6^{20}$ and $4^{20}$, respectively.

\textbf{Data processing and training setup}. We augment the ImageNet dataset before training, the specific operations including random cropping, light adjusting, and random horizontal flipping. During the training of the network, we implement the BN-based block indicator and BN-based layer indicator in search spaces based on the ShuffleNetV2 block and MobileNetV2 blocks, respectively. The epochs of the two sets of experiments are set to 80 and 100, and the learning rate is continuously adjusted with the growth of epochs using cosine annealing. In terms of search, we use an evolutionary algorithm to traverse each subnet according to the setting of the search strategy in SPOS\cite{guo2020single} to find the best subnet structure for retraining. Finally, for the searched subnet, 240 epochs of training are performed, and this process does not freeze any weights in the network. Our experiments are conducted on the NVIDIA GeForce RTX 3080 with the Pytorch framework.

\subsection{Comparison of Baseline Methods in the Search}

We compare the experimental results of the Supernet training phase in NAS with the results of baseline methods. Since our model only needs to optimize the parameters of the BN layer during back-propagation, the training time of our Supernet is dramatically reduced compared to other methods. In addition, due to the advantage of orthogonal weights, our network is much faster than BNNAS. The increase in FLOPS(468M) of the network after adding the BN-based layer indicator is minimal in exchange for a better performance of the searched network . The model's parameters stabilize in a lightweight range(4.9M) compared with other methods. Table~\ref{resultstable} shows the details of some models' size, computation cost, and training time. From the table it can be seen that any Supernet based algorithm has shorter running time than the general NAS algorithm(NASNet-A) and the BN-based indicator method(Ours, BNNAS) is an order of magnitude faster than the others.

\begin{table*}[t]
	\caption{Comparison of the results of the baseline method, some algorithms cannot be implemented on large datasets due to high computational cost, and then are experimented on the CIFAR dataset. During Supernet's training time we fully trained the parameters of BN block indicator and BN layer indicator with 80 epochs and 120 epochs separately. Accuracy and other figures in the brackets are reported by the original papers.}
	\label{resultstable}
	\renewcommand\arraystretch{1.2}
	\begin{center}
		\begin{tabular}{l|cccc}
			\toprule
			&\multicolumn{1}{c}{ \makecell[c]{FLOPS/Parameters\\(M/M)} } 
			&\multicolumn{1}{c}{ \makecell[c]{Acc.\\(\%)}}   
			&\multicolumn{1}{c}{\makecell[c]{Cost\\(GPU days)}}  
			&\multicolumn{1}{c}{Dataset}\\
			\midrule
			MobileNetV2(1.0)\cite{sandler2018mobilenetv2}   & 300/3.4  &72.0  &-   &-   \\
			ShuffleNetV2\cite{ma2018shufflenet}       & 286/3.7  &72.4  &-   &-   \\ 
			EfficientNet B0\cite{tan2019efficientnet}    & 390/5.3  &76.3  &-   &ImageNet  \\
			NASNet-A\cite{zoph2018learning}           & 564/5.3  &74.0  &2000 & CIFAR \\
			DARTS\cite{liu2018darts}              & 595/4.7  &73.1  &4.5 &CIFAR \\
			PC-DARTS\cite{xu2019pc}           & 597/5.3  &74.9  &3.7 &CIFAR \\
			GDARTS\cite{li2019gdarts}             & 497/4.4  &72.5  &-   &CIFAR \\
			Proxyless GPU\cite{cai2018proxylessnas}      & 465/7.1  &75.1  &8.3 &ImageNet \\
			One-Shot NAS\cite{bender2018understanding}       & -/5.1  &74.8  &4   &ImageNet \\
			SPOS\cite{guo2020single}               &323/3.5 &74.4 &11  &ImageNet \\
			FairNAS\cite{chu2021fairnas}            & 388/4.4  &74.7  &12  &ImageNet  \\
			BNNAS(FairNAS)\cite{chen2021bn}     & 326/3.7  &74.12  &1.2(15 epochs)  &ImageNet  \\
			BNNAS(SPOS)\cite{chen2021bn}       & 470/5.4  &75.67  &0.8(10 epochs)  &ImageNet  \\
			\midrule
			\textbf{Ours(without BN layer)}  & 468/4.9  & 75.76  & 2.8(80 epochs)  & ImageNet  \\
			\textbf{Ours(with BN layer)}    & 468/4.9  & 76.22  & 4.1(120 epochs)  & ImageNet \\
			\bottomrule
		\end{tabular}
	\end{center}
\end{table*}

There is a large gap between the method based on the BN indicator and other methods in training time , for the gradient propagation only focuses on the BN layer in the network. Compared to BNNAS, our network goes further in training time, and each image takes less propagation time in our network. Since we are not concerned with the expressiveness of the BN layers, train a sufficient number of epochs for the supernet. After 80 epoch iterations our supernet training time is 2.8 gpu days.

\subsection{Searching Approach}

As for the search algorithm, we follow BNNAS and SPOS, using an evolutionary algorithm that does not affect the time to search the structure even  adding the BN based layer indicator. We believe that our search algorithm with BN-based layer indicator can provide a better model structure with the addition of proxy requirements.

\subsection{Validation Set Evaluation}

Not only do we compare the results on the validation set, but also conduct comparative experiments with BNNAS for our design, including layer-based BN indicator and dynamic isometric network parameter initialization method. The search space of all models is implemented based on the MobileNetV2 block. We set up two sets of experiment models with orthogonal weight initialization, and both experiments use the BN block indicator. We added a BN-based layer indicator to one of the models. From the experimental results, it is also of benefit to the network’s search results to judge the essential indicators of each layer , and the effect of top-1 accuracy(76.22$\%$) is the best of all similar algorithms.

For the comparison of the initialization weight methods , we use different initialization methods for the initial model parameters and performed a complete forward and backward propagation. To be as fair as possible, we set the batchsize to 32 uniformly and calculate the latency of one complete training for each method. Other initialization methods we refer to open source code. In addition, it is difficult to control the specific initialization value of the model parameters, so we conduct several experiments to take the average value. 

Our method's latency(93.1 ms) narrowly outperforms BNNAS(96.3 ms).

\begin{table}[t]
	\caption{Compare the results with SPOS and BNNAS, in order to show the characteristics of our BN indicator and orthogonal initialization. For the Latency comparison we unify the experiments on the NVIDIA GeForce RTX 3080. In the table, `F' donates FairNAS, `S' donates SPOS.}
	\label{compareBNNASSPOS}
	\renewcommand\arraystretch{1.2}
	\begin{center}
		\begin{tabular}{l|ccc}
			\toprule
			&\multicolumn{1}{c}{\makecell[c]{Top-1 Acc. \\(\%)}}  
			&\multicolumn{1}{c}{\makecell[c]{Latency\\(ms)}} 
			&\multicolumn{1}{c}{\makecell[c]{Training \\Parameter}} \\ 
			\midrule
			SPOS   &74.4  &137.4  & ALL \\
			BNNAS(F)    &74.12   & -  &BN   \\
			BNNAS(S)  &75.67   &96.3   &BN  \\
			Ours(F)  &75.76   &-    &BN  \\
			Ours(F)   &76.22   &93.1    &BN (block+layer)  \\
			\bottomrule
		\end{tabular}
	\end{center}
\end{table}

In order to prove the generalization ability of the method, we change the search space to be based on the ShuffleNetV2 block module, and set the control of whether to add the BN-based layer indicator. As shown in table~\ref{compareBNNASSPOS}, Compared with the strategy of SPOS whose search space is also based on ShuffleNetV2 blocks modules, our method is faster and has fairer 
evaluation criteria for candidate modules. Our model is also more capable of finding the most efficient subnet structure. In addition, our model has the ability to  find multiple excellent models from a search space simultaneously.

\section{Conclusion}
NAS has greatly boosted the SOTA deep learning methods in computer vision. However, existing NAS methods are time-consuming and lack of fairness. We proposed a novel parameters initialization strategy for supernet pre-training which can efficiently ensure the fairness of subnets selection. The proposed method can also accelerate the searching process and improve the performance of searching results for NAS. Thanks to the mean field theory, we can guarantee that our method has theoretical support by analyzing the dynamics of random neural networks and estimating the generalization error of dynamic isometry block. Extensive experiments validate that the proposed dynamical isometry based rigorous Fair NAS can significantly decrease the whole time consumption for one-shot NAS without performance loss.

{\small
\bibliographystyle{ieee_fullname}
\bibliography{egbib}
}

\appendix

\section{Proof of Theorem 1}
To prove the theorem, we need the following definitions and lemmas:
\begin{definition} \label{definition1}
    The Orlicz norm of a random variable $X$ with respect to a convex function $\psi : [0, \infty) \rightarrow [0, \infty)$ 
    such that $\psi(0) = 0 $ and $\text{lim}_{x \rightarrow \infty} \psi(x) = \infty$ is defined by

    \begin{equation}
        \left\lVert X\right\rVert _{\psi} := \inf \left\{ t>0 | \mathbb{E} \left[ \psi \left( \frac{\left\lvert X\right\rvert }{t}\right)\right] \leq 1\right\} 
    \end{equation}

    if $\left\lVert X\right\rVert _{\psi_2} < \infty$ $X$ is called to be $sub-Gaussian$, and $sub-exponential$ if $\left\lVert X\right\rVert _{\psi_1} < \infty$ 
    where $\psi _{p}(x) := \exp \{ x^{p}\} - 1$ for $p \geq 1$.
\end{definition}

The following lemmas take advantage of the properties of the Orlicz norm and provide some inequalities for use as instructions for follow-up problems.

\begin{lemma} \label{two}
    Let $X$ and $Y$ be sub-Gaussian random variables. Then,

    \begin{enumerate}
        \item \textbf{Sum of independent sub-Gaussian.} If $X$ and $Y$ are also independent, then their sum, $X+Y$, 
            is sub-Gaussian. Moreover, $\left\lVert X+Y\right\rVert _{\psi_2}^2 \leq C\left( \left\lVert X\right\rVert _{\psi _2}^2 + \left\lVert Y\right\rVert _{\psi _2}^2\right)$ 
            for an absolute constant $C$. The same holds(with the same constant $C$) also for sums of multiple 
            independent sub-Gaussian random variables.
        \item \textbf{Centering.} $X-\mathbb{E}[X]$ is sub-Gaussian. Moreover, $\left\lVert X - \mathbb{E}[X]\right\rVert _{\psi _2} \leq C \left\lVert X\right\rVert _{\psi _2}$ 
            for an absolute constant $C$.
        \item \textbf{Product of sub-Gaussian.} $XY$ is sub-exponential. Moreover, $\left\lVert XY\right\rVert _{\psi _1} \leq \left\lVert X\right\rVert _{\psi _2} \left\lVert Y\right\rVert _{\psi _2}$
    \end{enumerate}
\end{lemma}

\begin{lemma}[Bernstein's inequality for sub-exponentials]\label{lam:ber}
    Let $X_1,...,X_N$ be independent zero-mean sub-exponential random variables. Then $\forall t \geq 0$:
    \begin{equation}
        \mathbb{P} \left( \left\lvert \frac{1}{N} \sum_{i=1}^{N} X_i \right\rvert \geq t \right) \leq 2 \exp \left\{ -\min \left\{ \frac{t^2}{K ^2}, \frac{t}{K}\right\} \cdot C \cdot N \right\},
    \end{equation}
    where $K= \text{max} _i \left\lVert X _i\right\rVert _{\psi _1}$ and $c > 0$ is an absolute constant.
\end{lemma}

We first fill the parameters of $F_{p}$ with random parameters that satisfy a Gaussian distribution $F_p\sim\mathcal{N}(0,v)$, then $\left\langle F_p,[x]_{ij}^r\right\rangle $ 
and $\left\langle F_p,[y]_{ij}^r\right\rangle $ are jointly Gaussian random variables with 
mean zero and variances $v ^2 \left\lVert [x]_{ij}^r\right\rVert $ and $v ^2 \left\lVert [y]_{ij}^r\right\rVert $ respectively, due to their independence and linear combinations. Hence, $\left\langle F_p,[x]_{ij}^r\right\rangle $ and $\left\langle F_p,[y]_{ij}^r\right\rangle $ are also sub-Gaussian with, $\forall p \in [N], \forall i,j \in \mathbb{Z}_n,$:
\begin{equation} \label{7}
    \begin{aligned}
        &\left\lVert \left\langle F_p,[x]_{ij}^r\right\rangle\right\rVert _{\psi _2} \leq C_0 v \left\lVert [x]_{ij}^r\right\rVert, \\
        &\left\lVert \left\langle F_p,[y]_{ij}^r\right\rangle\right\rVert _{\psi _2} \leq C_0 v \left\lVert [y]_{ij}^r\right\rVert, 
    \end{aligned}   
\end{equation}
where universal constant $C_0>0$. To achieve dynamic isometry, we rely on the conclusions of the previous orthogonal weights matrix obtained by triangulating every filter $F_p$. Suppose the $F_p$ is reversible, then there is a unique positive definite triangular matrix $W$ and its inverse $W^{-1}$, such that $Q_p = F_p \cdot W^{-1}$, where $Q_p \in \mathbb{R} ^{r \times r \times d}$ is an orthogonal matrix. We have:
\begin{equation}
   \begin{aligned}
      &\left\lVert \left\langle Q_p,[x]_{ij}^r\right\rangle\right\rVert _{\psi _2} \leq C_1 v \left\lVert [x]_{ij}^r\right\rVert, \\
      &\left\lVert \left\langle Q_p,[y]_{ij}^r\right\rangle\right\rVert _{\psi _2} \leq C_1 v \left\lVert [y]_{ij}^r\right\rVert,  \\
      &\forall p \in [N], \quad\forall i,j \in \mathbb{Z}_n, \quad C_1 = C_0 \cdot \left(\sum _{k=1}^{r} w_{k,k} ^2\right)^{-\frac{1}{2}}, \\
   \end{aligned} 
\end{equation}

where $w_{k,k}$ denote the elements on the diagonal of $W^{-1}$, $f_{kk}$ and $x_{ij_{kk}}$ in the following proof are also diagonal 
elements of $F_p$ and $[x]_{ij}^r$, respectively.

\begin{proof}[Proof of Theorem 1]
   \begin{equation}
      \begin{aligned}
         \left\lVert \left\langle Q_p,[x]_{ij}^r\right\rangle\right\rVert _{\psi _2} &= \left\lVert \text{tr}\left(F_p \cdot [x]_{ij}^r \cdot W ^{-1}\right)\right\rVert _{\psi _2} \\
         &\leq \left\lVert\sqrt{\sum _{k=1}^r \left(f_{kk} x_{ij_{kk}}\right)^2} \cdot \sqrt{\sum _{k=1}^{r} w_{k,k} ^2}\right\rVert _{\psi _2} \\
         &\leq \left\lVert\sum _{k=1}^r f_{kk} x_{ij_{kk}}\right\rVert _{\psi _2} \cdot \left\lVert\sqrt{\sum _{k=1}^{r} w_{k,k} ^2} \right\rVert _{\psi _2} \\
         &= \left\lVert \left\langle F_p,[x]_{ij}^r\right\rangle\right\rVert _{\psi _2} \cdot \left\lVert\sqrt{\sum _{k=1}^{r} w_{k,k} ^2} \right\rVert _{\psi _2}.
      \end{aligned}
   \end{equation}
    $C_0$ is a sufficiently large constant, and the magnitude of $C_0$ is much larger than the trace of the triangular matrix. Combining the above derivation results 
   with equation \ref{7}, we can get

   \begin{equation}
      \left\lVert \left\langle Q_p,[x]_{ij}^r\right\rangle\right\rVert _{\psi _2} \leq \left\lVert \left\langle Q_p,[x]_{ij}^r\right\rangle\right\rVert _{\psi _2} \leq C_1 v \left\lVert [x]_{ij}^r\right\rVert.
   \end{equation}

   where $C_1 =  \frac{C_0}{\sqrt{\left(\sum_{k=1}^{r} w_{k,k} ^2\right)}}$, for uniqueness of the triangular matrix $W$, $C_1$ is a constant.

The activation function is represented as $\sigma$, and we use the $tanh$ method here according to the previous reasoning, then

\begin{equation}
    X^{\prime}_{ij,p} := \sigma \left(\left\langle Q_p,[x]_{ij}^r\right\rangle \right), \quad Y^{\prime}_{ij,p} := \sigma \left(\left\langle F,[y]_{ij}^r\right\rangle \right), \quad
    \forall l \in [N],\quad \forall i,j \in \mathbb{Z}_n   
\end{equation}

Adding a batch normalization layer to the above filter calculation:
\begin{equation}
   \begin{aligned}
      X_{ij,p} &:= \sigma \left(\left\langle Q_p,[x]_{ij}^r\right\rangle \right) \cdot \frac{\gamma}{\sqrt{v_h-\hat{\varepsilon}}},\\ 
      Y_{ij,p} &:= \sigma \left(\left\langle F,[y]_{ij}^r\right\rangle \right)\cdot \frac{\gamma}{\sqrt{v_{h^{\prime}}-\hat{\varepsilon}}}, \\
      &\forall l \in [N],\quad \forall i,j \in \mathbb{Z}_n  
   \end{aligned}
\end{equation}

Set $\sigma$ is Lipschitz continuous with a Lipschitz constant $L$. In addition, since each module satisfies the dynamic isometry, the input and output variance is constant then the parameters of the BN layer are independent of others. We can get $X_{ij,p}$ and $Y_{ij,p}$ are sub-Gaussian with:
\begin{equation}
   \left\lVert X_{ij,p}\right\rVert _{\psi _2} \leq C_1 \left\lVert \gamma\right\rVert  Lv \left\lVert [x]_{ij}^r\right\rVert v_h^{\prime}, \text{  } 
   \left\lVert Y_{ij,p}\right\rVert _{\psi _2} \leq C_1 \left\lVert \gamma\right\rVert  Lv \left\lVert [y]_{ij}^r\right\rVert v_{h^{\prime}}^{\prime},\quad 
   \forall p \in [N], \text{  } \forall i,j \in \mathbb{Z}_n
\end{equation}
where $v_{h}^{\prime} = \left\lVert (v_h - \hat{\varepsilon})^{-\frac{1}{2}}\right\rVert $, $v_{h^{\prime}}^{\prime} = \left\lVert (v_{h^{\prime}} - \hat{\varepsilon})^{-\frac{1}{2}}\right\rVert $ $\gamma$ are trainable parameters.

Therefore, by the lemma \ref{two} we continue to have:
\begin{equation}
    \begin{aligned}
        \left\lVert X_{ij,p}Y_{ij,p}\right\rVert _{\psi_1} &\leq C_1^2 \left\lVert \gamma\right\rVert  ^2 v^2 L^2 \left\lVert [x]_{ij}^r\right\rVert\left\lVert [y]_{ij}^r\right\rVert v_{h^{\prime}}^{\prime} v_h^{\prime}, \\
        \left\lVert X_{ij,p}Y_{ij,p} - \mathbb{E}[X_{ij,p}Y_{ij,p}]\right\rVert _{\psi_1} &\leq C_1^2 C \left\lVert \gamma\right\rVert ^2 v^2 L^2 \left\lVert [x]_{ij}^r\right\rVert\left\lVert [y]_{ij}^r\right\rVert v_{h^{\prime}}^{\prime} v_h^{\prime}
    \end{aligned}
\end{equation}

For $[x]_{ij}^r$ and $[y]_{ij}^r$ at all locations in the input signals, we let their minimum maximum related to $R$, then bring in the distance formula we want to find:
$$
\mathbb{P} \left(\left\lvert \frac{1}{N} \sum_{p=1}^{N} \left\langle B\left( \sigma\left( F_p * \mathbf{h}\right)\right),B\left( \sigma\left( F_p * \mathbf{h}^{\prime}\right)\right)\right\rangle -  
      \mathbb{E} \left[ \left\langle B\left( \sigma\left( F * \mathbf{h}\right)\right), B\left( \sigma\left( F * \mathbf{h}^{\prime}\right)\right)\right\rangle \right]\right\rvert \geq \varepsilon \right) 
$$

\begin{equation}
   \begin{aligned}
      &= \mathbb{P} \left( \frac{1}{N} \left\lvert \sum_{p\in [N]; i,j \in \mathbb{Z}_n} \left\{ X_{ij,p}Y_{ij,p}-\mathbb{E}[X_{ij,p}Y_{ij,p}]\right\} \right\rvert \geq \varepsilon\right) \\
      &\leq \mathbb{P} \left( \frac{1}{N} \sum_{i,j \in \mathbb{Z}_n}  \left\lvert \sum_{p \in [N]}\left\{ X_{ij,p}Y_{ij,p}-\mathbb{E}[X_{ij,p}Y_{ij,p}]\right\}\right\rvert  \geq \varepsilon\right) \\
      &\leq \mathbb{P} \left( \frac{n^2}{N} \max_{i,j \in \mathbb{Z}_n} \left\lvert \sum _{p\in [N]} \left\{ X_{ij,p}Y_{ij,p}-\mathbb{E}[X_{ij,p}Y_{ij,p}]\right\}\right\rvert \geq \varepsilon\right) \\
      &\leq \sum_{i,j\in \mathbb{Z}_n} \mathbb{P} \left( \frac{n^2}{N} \left\lvert \sum_{p \in [N]}\left\{ X_{ij,p}Y_{ij,p}-\mathbb{E}[X_{ij,p}Y_{ij,p}]\right\}\right\rvert  \geq \varepsilon\right) \\
      &\overset{\text{lemma}\ref{lam:ber}}{\leq} 2n^2 \exp \left\{ -min \left(\frac{\varepsilon ^2}{C_1^4 C^2 \left\lVert \gamma\right\rVert ^4 v^4 L^4 R^4 n^4}, \frac{\varepsilon}{C_1^2 C \left\lVert \gamma\right\rVert ^2 v^2 L^2 R^2 n^2}\right)cN\right\} = \delta.
   \end{aligned}
\end{equation}
\end{proof}

\end{document}